\documentclass[submission]{eptcs}
\providecommand{\event}{TARK 2025} 
\usepackage{breakurl}             
\usepackage{underscore}           

\newtheorem{THEOREM}{Theorem}[section]
\newenvironment{theorem}{\begin{THEOREM} \hspace{-.85em} {\bf :} }%
                        {\end{THEOREM}}
\newtheorem{LEMMA}[THEOREM]{Lemma}
\newenvironment{lemma}{\begin{LEMMA} \hspace{-.85em} {\bf :} }%
                      {\end{LEMMA}}
\newtheorem{COROLLARY}[THEOREM]{Corollary}
\newenvironment{corollary}{\begin{COROLLARY} \hspace{-.85em} {\bf :} }%
                          {\end{COROLLARY}}
\newtheorem{PROPOSITION}[THEOREM]{Proposition}
\newenvironment{proposition}{\begin{PROPOSITION} \hspace{-.85em} {\bf :} }%
                            {\end{PROPOSITION}}
\newtheorem{DEFINITION}[THEOREM]{Definition}
\newenvironment{definition}{\begin{DEFINITION} \hspace{-.85em} {\bf :} \rm}%
                            {\end{DEFINITION}}
\newtheorem{CLAIM}[THEOREM]{Claim}
\newenvironment{claim}{\begin{CLAIM} \hspace{-.85em} {\bf :} \rm}%
                            {\end{CLAIM}}
\newtheorem{EXAMPLE}[THEOREM]{Example}
\newenvironment{example}{\begin{EXAMPLE} \hspace{-.85em} {\bf :} \rm}%
                            {\end{EXAMPLE}}
\newtheorem{REMARK}[THEOREM]{Remark}
\newenvironment{remark}{\begin{REMARK} \hspace{-.85em} {\bf :} \rm}%
                            {\end{REMARK}}

\newcommand{\thm}{\begin{theorem}}
\newcommand{\lem}{\begin{lemma}}
\newcommand{\pro}{\begin{proposition}}
\newcommand{\dfn}{\begin{definition}}
\newcommand{\rem}{\begin{remark}}
\newcommand{\xam}{\begin{example}}
\newcommand{\cor}{\begin{corollary}}
\newcommand{\prf}{\noindent{\bf Proof:} }
\newcommand{\ethm}{\end{theorem}}
\newcommand{\elem}{\end{lemma}}
\newcommand{\epro}{\end{proposition}}
\newcommand{\edfn}{\bbox\end{definition}}
\newcommand{\erem}{\bbox\end{remark}}
\newcommand{\exam}{\bbox\end{example}}
\newcommand{\ecor}{\end{corollary}}
\newcommand{\eprf}{\bbox\vspace{0.1in}}
\newcommand{\beqn}{\begin{equation}}
\newcommand{\eeqn}{\end{equation}}

\newcommand{\bbox}{\vrule height7pt width4pt depth1pt}

\newcommand{\clm}{\begin{claim}}
\newcommand{\eclm}{\end{claim}}

\newcommand{\sat}{\models}

\newcommand{\rimp}{\Rightarrow}

\newcommand{\dimp}{\Leftrightarrow}
\newcommand{\bor}{\bigvee}

\newcommand{\union}{\cup}

\renewcommand{\phi}{\varphi}

\newcommand{\C}{{\cal C}}

\newcommand{\F}{{\cal F}}

\newcommand{\K}{{\cal K}}
\newcommand{\M}{{\cal M}}

\newcommand{\R}{{\cal R}}

\newcommand{\U}{{\cal U}}
\newcommand{\V}{{\cal V}}
\newcommand{\W}{{\cal W}}
\newcommand{\X}{{\cal X}}

\newcommand{\ol}{\setlength{\itemsep}{0pt}\begin{enumerate}}
\newcommand{\eol}{\end{enumerate}\setlength{\itemsep}{-\parsep}}
\newcommand{\ul}{\setlength{\itemsep}{0pt}\begin{itemize}}
\newcommand{\dl}{\setlength{\itemsep}{0pt}\begin{description}}
\newcommand{\edl}{\end{description}\setlength{\itemsep}{-\parsep}}
\newcommand{\eul}{\end{itemize}\setlength{\itemsep}{-\parsep}}

\newcommand{\BS}{B^{\scriptscriptstyle \cS}}

\newcommand{\false}{{\it false}}

\newcommand{\commentout}[1]{}

\newcommand{\bi}{\begin{itemize}}
\newcommand{\ei}{\end{itemize}}
\newcommand{\be}{\begin{enumerate}}
\newcommand{\ee}{\end{enumerate}}

\newenvironment{oldthm}[1]{\par\noindent{\bf Theorem #1:} \em \noindent}{\par}
\newenvironment{oldlem}[1]{\par\noindent{\bf Lemma #1:} \em \noindent}{\par}
\newenvironment{oldcor}[1]{\par\noindent{\bf Corollary #1:} \em \noindent}{\par}
\newenvironment{oldpro}[1]{\par\noindent{\bf Proposition #1:} \em \noindent}{\par}
\newcommand{\othm}[1]{\begin{oldthm}{\ref{#1}}}
\newcommand{\eothm}{\end{oldthm} \medskip}
\newcommand{\olem}[1]{\begin{oldlem}{\ref{#1}}}
\newcommand{\eolem}{\end{oldlem} \medskip}
\newcommand{\ocor}[1]{\begin{oldcor}{\ref{#1}}}
\newcommand{\eocor}{\end{oldcor} \medskip}
\newcommand{\opro}[1]{\begin{oldpro}{\ref{#1}}}
\newcommand{\eopro}{\end{oldpro} \medskip}

\newcommand{\bxor}[1]{\dot{\bor}}

\renewcommand{\eothm}{\end{oldthm}}

\usepackage{graphicx} 
\renewcommand{\W}{S}
\newcommand{\Cond}{\rightarrow}
\newcommand{\Scal}{{\cal S}}
\newcommand{\ST}{\mbox{{\it ST}}}
\newcommand{\BH}{\mbox{{\it BH}}}
\newcommand{\BT}{\mbox{{\it BT}}}
\renewcommand{\BS}{\mbox{{\it BS}}}
\newcommand{\SH}{\mbox{{\it SH}}}
\newcommand{\Lex}{{\cal L}^+}
\newcommand{\LM}{{\cal L}^{\M}}
\renewcommand{\L}{{\cal L}}
\newcommand{\nciteyear}{\cite}
\newcommand{\citeyear}{\cite}
  \newcommand{\fullv}[1]{}
\newcommand{\shortv}[1]{#1}

\title{Causality Without Causal Models}
\author{Joseph Y. Halpern
\institute{Cornell University}
\email{halpern@cs.cornell.edu}
\and
Rafael Pass
\institute{Cornell University}
\email{rafael@cs.cornell.edu}
}

\def\titlerunning{Causality Without Causal Models}
\def\authorrunning{J.Y. Halpern \& R. Pass}

\begin{document}

\maketitle

\begin{abstract}
  Perhaps the most prominent current definition of (actual) causality is due to
  Halpern and Pearl.  It is defined using causal models (also known as
\emph{structural equations models}).  We abstract the definition,
extracting its
key features, so that it can be applied to any other model where
counterfactuals are defined.
By abstracting the definition, we gain a number of benefits. Not only
can we apply the definition in a wider range of models, including ones
that allow, for example, backtracking, but we
can
apply the definition
to determine if $A$ is a cause of $B$  even if $A$
and $B$ are formulas involving disjunctions, negations, beliefs, and
nested counterfactuals (none of which can be handled by the
Halpern-Pearl definition).
Moreover, we can extend the ideas to getting an abstract definition of
explanation that can be applied beyond causal models.
Finally, we gain a
deeper understanding of features of the definition even in causal
models. 
\end{abstract}

\vspace{-0.1in}
\section{Introduction}
  Perhaps the most prominent current definition of (actual) causality
  is due to 
  Halpern and Pearl \cite{Hal48,HP01b} (HP from now on).  It is
  defined using causal 
  models (also known as 
  \emph{structural equations models}).  The fact that the definition
  is so tied to causal models has benefits---for example, the fact that causal
  models can be represented graphically makes them relatively easy to
  work with---but also leads to a number of limitations; specifically,
  the language in which we can express counterfactuals is limited
  (so we cannot express causal statements that involve more complex
  counterfactuals), and we cannot 
  deal with \emph{backtracking}.  (See below for further discussion.)
  The goal of this paper is to abstract the definition of
  causality in causal models, extracting its
key features, so that it can be applied to any other model where
counterfactuals are defined.

Doing so lets us get around these limitations.  For one thing, it lets
us consider more 
general causal statements. 
The HP definition of ``$A$ is a cause of $B$'' in causal models
can be applied only when
$A$ is a conjunction of primitive events $X=x$, where
$X$ is a variable and $x$ is a value in the range of $X$, and $B$ is
a Boolean combination of primitive events.  It
cannot be applied if $A$ is
a disjunction, or if either $A$ or $B$ involves 
  modalities like belief or nested counterfactuals.
But such
  statements are important in practice!  For example, Sartorio
  \nciteyear{Sartorio06} gives examples where disjunctive causes play a
  significant role; Soloviev and Halpern \nciteyear{SH21} give examples
  where nested counterfactuals are critical for expressing security
  properties.  And we clearly want to be able to reason about 
 (and intervene on) agent's beliefs. For example, we might want to
  say ``If were to intervene on Alice's beliefs so that she believed
  that the vaccine was effective, then 
    she would take it''.

  Causal models have been extended to allow counterfactuals
that involve disjunctions and nested counterfactuals \cite{Briggs12},
and reasoning about both knowledge/belief and causality \cite{BSVX21}
(although not interventions on beliefs).  These extensions involving
disjunction and nested counterfactuals are not at all
straightforward, and lead to some arguably unreasonable axioms. For
example, the axiom given by Briggs that characterizes interventions
involving disjunctions is $[A   \lor A']B \dimp ([A]B \land [A']B
\land [A\land A']B$).  It is not clear why this is appropriate.
Briggs does not try to justify it and, indeed, admits that the
axiom validates ``the controversial 
rule \emph{Simplification of Disjunctive Antecedents} \ldots [which]
most authors reject''.  Dealing with interventions on beliefs also
seems quite nontrivial.   All this means that a definition of
causality involving these constructs may well be controversial.  The
advantage of the approach that we are suggesting here is that,
as long
as we start with a framework that has a definition of counterfactuals
that a user is 
comfortable with and supports these constructs,
we can essentially import the HP definition to that framework.  
In particular, we get a definition of causality in
the counterfactual structures considered by Stalnaker
\nciteyear{Stalnaker68} and Lewis \nciteyear{Lewis73} that are
standard in the philosophy literature.  Moreover, we can show that this
definition 
generalizes that in causal
models: as shown by Halpern~\nciteyear{Hal40},
causal models correspond to a subset of counterfactuals; we show here
(see Theorem~\ref{pro:nobacktracking}) that, in a causal model 
$M$, the original definition of causality agrees with the new
definition
in the counterfactual structures $M'$ corresponding to $M$.

As we said above, the second key limitation of defining causality in
causal models is that we cannot  deal with backtracking.
As has often been observed, when considering the effect of an
intervention such as $X=x$ in a causal model, no variables
``upstream'' of $X$ in the causal graph 
are affected; only
descendants of $X$ can be affected.  This is referred to as ``no
backtracking''.

The argument for no backtracking is not hard to explain.  If we
intervene to set, say $X=2$, then, intuitively, all that should
change are the values of variables that are descendants of $X$, since
these are the only ones that can be affected by the intervention.
But, as argued by von K\"ugelen et al.~\nciteyear{KMB23}, there are
cases where we want to ascribe causality but do not want to consider
counterfactuals that result from interventions.  Intuitively, these are cases
where the causal laws, not the background conditions, are shared between
the actual and counterfactual worlds. Consequently, the ``upstream''
variables must be allowed to 
differ to accommodate possibly contradictory facts.
As discussed by von K\"ugelen et al. (who also give references to work
in the cognitive psychology literature on the topic), people certainly
consider 
backtracking counterfactuals when evaluating causality.  This makes it
useful to have a definition of causality that allows them.
Causal models have been extended so as to deal with backtracking
\cite{KMB23,LK15}, but the definitions of causality used are quite
different from the HP definition.  Our
approach lets us import the HP definition ``for free'', so to speak,
as long as we 
start with a definition of counterfactual that allows backtracking
(again, as do the standard definitions of Stalnaker
\nciteyear{Stalnaker68} and Lewis \nciteyear{Lewis73}).

This approach can be further extended to give an abstract definition
of \emph{explanation}.  There are many definitions of explanations in
the literature.  The modern literature goes back to the work of Hempel
\nciteyear{Hempel65} and Salmon \nciteyear{Salmon70}, but this work is
well known to have problems because it does not take causality into
account.  Here we focus on causal explanations, again taking as our
basis the work of Halpern and Pearl \cite{Hal48,HP01a}.  The
definition is based on the defintion of causality (which is what will
allow us to apply the approach of this paper), but it also takes into
account
the well-known observation
that what counts as an explanation is relative to what an
agent knows \cite{Gardenfors1,Salmon84}.
As G\"ardenfors \nciteyear{Gardenfors1} observes, an agent seeking 
an explanation of why Mr.\ Johansson has been taken ill with lung cancer
will not consider the fact that he worked for years in asbestos
manufacturing an explanation if he already knew this fact.
It is relatively straightforward to take an agent's knowledge into
account in our abstract approach.  Doing so gives us the tools we need
to handle explanation as well.


\paragraph{Our Definition in a Nutshell}
The key intuition for the definition is the
classic
notion of \emph{but-for}
causality used in the law
literature: $A$ is a cause of $B$ if, but for $A$, $B$
would not have happened.  As is well known (see
Example~\ref{xam:rt}), but-for causality does not suffice to
give a definition of causality.  There are times we would like to call
$A$ a cause of $B$ even if $A$ is not a but-for cause of $B$.   The
definition of actual causality in  causal  models and our new 
abstract definition can both be viewed as essentially saying that $A$
is a cause of $B$ if, conditional on some appropriate $C$, $A$ is a
but-for cause of $B$.
Like the definition of \cite{Hal48},
we just require $C$ to be a formula that is true (in the
current state of the world).

In more detail, our
\fullv{notion, which we refer to as \emph{conditional
  but-for-causality},}
\shortv{notion} says that \emph{$A$ is an actual cause of $B$}
with respect to a language $\L$ if, roughly speaking, the following
three conditions (which are formalized in
Section~\ref{sec:abstraction}) hold:
\begin{description}
\item[{\rm AC1$'$.}] $A$ and $B$ both hold (in the true state of the world);
\item[{\rm AC2$'$.}] There is a formula $C \in \L$ such that $C$ holds
  (in the true state of the world), and if $\neg A \land C$ were to
  hold (counterfactually) then $\neg B$ would hold.
\item[{\rm AC3$'$.}] $A$ is minimal; that is, there is no formula $A'$
  that implies $A$ yet is not implied by $A$ that also satisfies AC2$'$.
\end{description}
\commentout{
To formalize this notion, we just need to have a way to interpret
counterfactuals, which can be done in a number of standard ways.
In fact, this definition makes sense even if we have only an
informal way of interpreting counterfactuals (which may be
useful, e.g., in legal contexts).}

Note that the  definition above is parametrized by a
language $\L$. As we show, in case this language $\L$
consists only of conjunctions of primitive events (and
$A,B$ are also
restricted to this language), then our definition collapses down to the
HP definition of causality (see
Theorem~\ref{thm:defsagree}). Furthermore, if the language is
slightly richer, then the same is true even in \emph{recursive
counterfactual structures}, counterfactual structures that essentially
capture causal models (see Theorem~\ref{pro:nobacktracking}).

\paragraph{Paper Outline}
The rest of the paper is organized as follows.  In
Section~\ref{sec:review},
we review the HP definition of actual causality, as modified by
Halpern~\nciteyear{Hal48}.  In Section~\ref{sec:ccf}, we discuss
\emph{causal-counterfactual families (of models)---ccfs}, a general class
of frameworks to which our abstract definition applies. We show that
causal models and the counterfactual structures considered in
the philosophy literature are instances of ccfs.
In
Section~\ref{sec:abstraction}, we give 
our abstract definition of causality, show that the HP definition is a
special case of it, and consider its implications in counterfactual
structures.  We consider backtracking in
Section~\ref{sec:backtracking},
and explanation in Section~\ref{sec:explanation}.
Proofs are deferred to the appendix.

\vspace{-0.1in}
\section{Actual causality in causal models}\label{sec:review}

Here we briefly review the definition of causal models
introduced by Halpern and Pearl \nciteyear{HP01b}, and the definition of
actual causality in causal
models.  In fact, three variants of the definition have been proposed,
called the ``original'', ``updated'', and ``modified'' definitions
\cite{Hal48}.  We focus here on the modified definition, both because
it is the simplest to state and, as shown by Halpern \nciteyear{Hal48}, the
easiest to work with, and because it lends itself naturally to
generalization.
The material in the first three subsections of this section is largely
taken from \cite{Hal48}.
While there are a number of other definitions of causality in causal
models (e.g., \cite{GW07,Hall07,hitchcock:99,Hitchcock07,Woodward03}),
we focus on the HP definition here because our generalization is based 
on it and because it has been the most influential (judging by Google scholar
citations).   We believe that our approach should be
applicable to other ways of defining causality, but we have not
checked details.

\vspace{-0.1in}
\subsection{Causal models}
We assume that the state is described in terms of 
variables and their values.  
Some variables may have a causal influence on others. This
influence is modeled by a set of {\em structural equations}.
It is conceptually useful to split the variables into two
sets: the {\em exogenous\/} variables, whose values are
determined by 
factors outside the model, and the
{\em endogenous\/} variables, whose values are ultimately determined by
the exogenous variables.  
The structural equations
describe how the latter values are 
determined.

Formally, a \emph{causal model} $M$
is a pair $(\Scal,\F)$, where $\Scal$ is a \emph{signature}, which explicitly
lists the endogenous and exogenous variables  and characterizes
their possible values, and $\F$ defines a set of \emph{(modifiable)
structural equations}, relating the values of the variables.  
A signature $\Scal$ is a tuple $(\U,\V,\R)$, where $\U$ is a set of
exogenous variables, $\V$ is a set 
of endogenous variables, and $\R$ associates with every variable $X \in 
\U \union \V$ a nonempty set $\R(X)$ of possible values for 
$X$ (i.e., the set of values over which $X$ {\em ranges}).  
For simplicity, we assume here that $\V$ is finite, as is $\R(X)$ for
every endogenous variable $X \in \V$.
$\F$ associates with each endogenous variable $X \in \V$ a
function denoted $F_X$
(i.e., $F_X = \F(X)$)
such that $F_X: (\times_{U \in \U} \R(U))
\times (\times_{Y \in \V - \{X\}} \R(Y)) \rightarrow \R(X)$.
This mathematical notation just makes precise the fact that 
$F_X$ determines the value of $X$,
given the values of all the other variables in $\U \union \V$.
If there is one exogenous variable $U$ and three endogenous
variables, $X$, $Y$, and $Z$, then $F_X$ defines the values of $X$ in
terms of the values of $Y$, $Z$, and $U$.  For example, we might have 
$F_X(u,y,z) = u+y$, which is usually written as
$X = U+Y$.   Thus, if $Y = 3$ and $U = 2$, then
$X=5$, regardless of how $Z$ is set.%

The structural equations define what happens in the presence of external
interventions. 
Setting the value of some variables $\vec{X}$ to $\vec{x}$ in a causal
model $M = (\Scal,\F)$ results in a new causal model, denoted
$M_{\vec{X}\gets \vec{x}}$, which is identical to $M$, except that,
for each $X \in \vec{X}$, the
equation in $\F$  for $X$ is replaced by $X = x$, where
$x$ is the component of $X$ in $\vec{x}$.

The dependencies between variables in a causal model $M = ((\U,\V,\R),\F)$
can be described using a {\em causal network}\index{causal
  network} (or \emph{causal graph}),
whose nodes are labeled by the endogenous and exogenous variables in
$M$, with one node for each variable in $\U \cup
\V$.  The roots of the graph are (labeled by)
the exogenous variables.  There is a directed edge from  variable $X$
to $Y$ if $Y$ \emph{depends on} $X$; this is the case
if there is some setting of all the variables in 
$\U \union \V$ other than $X$ and $Y$ such that varying the value of
$X$ in that setting results in a variation in the value of $Y$; that
is, there is 
a setting $\vec{z}$ of the variables other than $X$ and $Y$ and values
$x$ and $x'$ of $X$ such that
$F_Y(x,\vec{z}) \ne F_Y(x',\vec{z})$.
A causal model  $M$ is \emph{recursive} (or \emph{acyclic})
if its causal graph is acyclic.
If $M$ is an acyclic  causal model,
then given a \emph{context}, that is, a setting $\vec{u}$ for the
exogenous variables in $\U$, the values of all the other variables are
determined (i.e., there is a unique solution to all the equations).
We can determine these values by starting at the top of the graph and
working our way down.
In this paper, following most of the literature, we restrict to recursive models.

\vspace{-0.1in}
\subsection{Reasoning about causality}

Given a signature $\Scal= (\U,\V,\R)$, a \emph{primitive event} is a
formula of the form $X = x$, for  $X \in \V$ and $x \in \R(X)$.  
A {\em basic causal formula (over $\Scal$)\/} is one of the form
$[Y_1 \gets y_1, \ldots, Y_k \gets y_k] \varphi$,
where
$\varphi$ is a Boolean
combination of primitive events,
$Y_1, \ldots, Y_k$ are distinct variables in $\V$, and
$y_i \in \R(Y_i)$.
Such a formula is abbreviated
as $[\vec{Y} \gets \vec{y}]\varphi$.
The special
case where $k=0$
is abbreviated as
$\varphi$.
Intuitively,
$[Y_1 \gets y_1, \ldots, Y_k \gets y_k] \varphi$ says that
$\varphi$ would hold if
$Y_i$ were set to $y_i$, for $i = 1,\ldots,k$.
A {\em causal formula\/} is a Boolean combination of basic causal formulas.
Let $\L(\Scal)$ consist of all causal formulas over the signature
$\Scal$. (We typically omit $\Scal$ when it is clear from context.)

A pair $(M,\vec{u})$ consisting of a causal model $M$ and a
context $\vec{u}$ is a \emph{(causal) setting}.
A causal formula $\psi$ is true or false in a setting.
We write $(M,\vec{u}) \models \psi$  if
the causal formula $\psi$ is true in
the setting $(M,\vec{u})$.
The $\models$ relation is defined inductively.
$(M,\vec{u}) \models X = x$ if
the variable $X$ has value $x$
in the unique (since we are dealing with acyclic models) solution
to the equations in
$M$ in context $\vec{u}$ (i.e., the
unique vector of values for the variables that simultaneously satisfies all
equations 
in $M$ 
with the variables in $\U$ set to $\vec{u}$);
$(M,\vec{u}) \models [\vec{Y} \gets \vec{y}]\varphi$ if 
$(M_{\vec{Y} = \vec{y}},\vec{u}) \models \varphi$;
and Boolean combinations are defined in the standard way.

\vspace{-0.1in}
\subsection{The modified definition of actual causality}

We are now in a position to state the (modified) definition of actual
causality in causal models.

\dfn\label{actcaus}
$\vec{X} = \vec{x}$ is an \emph{actual cause of $\phi$ in
the causal setting $(M, \vec{u})$}
if the following three conditions hold:\index{causal setting}
\begin{description}
\item[{\rm AC1.}]\label{ac1} $(M,\vec{u}) \sat (\vec{X} = \vec{x})$ and 
$(M,\vec{u}) \sat \phi$.\index{AC1}
\item[{\rm AC2.}]\label{ac2} 
  There is a set $\vec{W}$ of variables in $\V$
  disjoint from $\vec{X}$,
$\vec{w}^* \in \R(\vec{W})$, and $\vec{x} \in \R(\vec{X})$
such that $(M,\vec{u})  \sat \vec{W} = \vec{w}^*$ and
$(M,\vec{u}) \sat [\vec{X} \gets \vec{x}', \vec{W} \gets
  \vec{w}^*]\neg \phi$.%
\protect{\footnote{Although the definition in \cite{Hal48} does not explicitly
  say that $\vec{W}$ is disjoint from $\vec{X}$, since interventions
  are performed on distinct variables, this follows from the
  notation.}}

\item[{\rm AC3.}] \label{ac3}\index{AC3}  
  $\vec{X}$ is minimal; there is no strict subset $\vec{X}'$ of
  $\vec{X}$ such that $\vec{X}' = \vec{x}'$ satisfies
condition AC2, where $\vec{x}'$ is the restriction of
$\vec{x}$ to the variables in $\vec{X}$.%
\footnote{Note that $\vec{X}' = \vec{x}'$ is guaranteed to satisfy
AC1, which is why only AC2 is mentioned here.}
\end{description}
\edfn

\vspace{-0.1in}
\subsection{The role of AC2}

The motivation for AC2 was to be able to deal with examples like the
following, due to Lewis \nciteyear{Lewis00}.  (The following
analysis is largely taken from \cite{Hal48}.)

\begin{example}\label{xam:rt}
Suzy and Billy both pick up rocks
and throw them at  a bottle.
Suzy's rock gets there first, shattering the
bottle.  Because both throws are perfectly accurate, Billy's would have
shattered the bottle
had it not been preempted by Suzy's throw.

We would like to call Suzy's throw the cause of the bottle
shattering.  This is the case in an appropriate causal model.  The
following causal model $M^{rt}$ does the job.
There are five endogenous variables:
\begin{itemize}
\item $\ST$ for ``Suzy throws'', with values 0 (Suzy does not throw) and
1 (she does);
\item $\BT$ for ``Billy throws'', with values 0 (he doesn't) and
1 (he does);
\item $\BS$ for ``bottle shatters'', with values 0 (it doesn't shatter)
and 1 (it does).
\item $\BH$ for ``Billy's rock hits the (intact) bottle'', with values 0
(it doesn't) and 1 (it does); and
\item $\SH$ for ``Suzy's rock hits the bottle'', again with values 0 and
1.
\end{itemize}
For simplicity, assume that there is one exogenous
variable $U$, with values $ij$, $i, j
\in \{0,1\}$, which   determines whether Billy and Suzy throw.  

We consider the following equations for the variables:
\begin{itemize}
\item $\ST =1$ iff $U \in \{10,11\}$;
  \item $\BT =1$ iff $U \in \{01,11\}$;
\item $\SH=1$ iff $\ST=1$; 
\item $\BH = 1$ iff $\BT = 1$ and $\SH = 0$ (this builds into the
  equation the fact that Billy rock does not hit the intact bottle if
  Suzy's rock already hit it);
  \item $\BS=1$ iff $\SH=1$ or $\BH= 1$.
\end{itemize}

Now consider the context $U=11$, where both Suzy and Billy throw rocks.
In this context, most people want to view $\ST=1$ as a cause of the
bottle shattering, while $\BT=1$ 
is not. But $\ST=1$ is not a but-for cause of the bottle shattering;
had Suzy not thrown, the bottle would have shattered anyway (Billy's
rock would have hit it).
Nevertheless, it is a cause.  
To show that $\ST=1$ is a cause, we take $W$ in AC2 to be
$\{BH\}$, and fix $\BH$ at its actual value of 0 (in fact, Billy
's throw does not hit the bottle).    Note that $(M,11) \models
[\ST\gets 0,\BH\gets 0]\BS=0$, so AC2 holds for $\ST=0$; clearly AC1 and AC3
hold as well.  Thus, $\ST=1$ is a cause of $\BS=1$ in the context
$U=11$.  It is easy to see 
that the symmetric argument does \emph{not} show that $\BT=1$ is a
cause.  We cannot fix $\SH=0$, because in the actual state, $\SH$ is
1, not 0.
We can think of $\ST=1$ as a but-for cause of $\BS=1$, conditional on
$\BH=0$.
\bbox
\end{example}

The idea of fixing some variables at their actual values, as embodied
by AC2, seems at first somewhat \emph{ad hoc}.  While it does give us
the ``right'' answer in the example above, why should we want to do it
in general?  The idea of keeping some variables fixed at their actual
value turns out to be quite a powerful one.  This is perhaps best seen
if we consider causality along a path.  To take a simple example,
suppose that someone's gender affects the outcome of a loan decision in
two ways.  First, gender affects salary, which in turns affects the
decision.  Second, gender affects how reliable someone is viewed as
being (say, for example, that
men are viewed as less likely to repay loans, even fixing everything
else).  That is, we have a simple causal graph with variables $G$
(gender) that affects both $S$ (salary) and $R$ (reliability), while
$S$ and $R$ both affect $L$ (loan repayment).  

We might decide that it is legitimate to take salary into
account when considering loan repayment, but not the perception of
reliability.  That is, we might decide that it is legitimate to
consider the effect of gender along the path $G-S-L$, but not along
the path $G-R-L$.   We can model causality along a path by simply
fixing the values of variables at their actual values off the path.
In this case, since we do not want to consider the impact of
perception of reliability, we fix $R$ at its actual value, which
allows us to consider the impact of changing $G$ along the path
$G-S-L$.%
\footnote{Thanks to Sander  Beckers for pointing this out to us.}

Fixing the values of some variables $\vec{W}$ is useful beyond being
able to capture path causality.
It allows us to consider causality
conditional on certain features of the state being kept fixed---the
key feature of our more abstract generalized  definition. 

\vspace{-0.1in}
\section{Causal-counterfactual families}\label{sec:ccf}

We want to define causality for all families of structures that
satisfy certain minimal conditions.  To that end, we consider (for
lack of a better name) \emph{causal-counterfactual families (ccfs)}.
A ccf $\M$ is a family of models or structures (we use the latter two
words more or less interchangeably) such that each model $M \in \M$
has a set of possible \emph{states} or worlds, and there is a
semantics that allows us to define $(M,w) \sat \phi$ for all formulas
in a language $\LM$.  We require that $\LM$ include primitive events in
some signature $\Scal$, and be closed under conjunction and negation
(so contains $\L(\Scal)$) and \emph{basic counterfactuals}, so that
if $\phi$ and 
$\psi$ are in $\L(\Scal)$, then $\phi \Cond \psi \in \LM$.  But $\LM$ may also 
be far more expressive and include, for example, nested
counterfactuals, and modalities such as belief and time.
The richness of $\LM$ is exactly what will allow our abstract definition of
causality to apply to quite expressive languages.

\vspace{-0.1in}
\subsection{Causal models as a ccf}
We can almost view causal models (over $\Scal$) as a ccf $\M$.  The
states are the  contexts. Let the language $\Lex(\Scal)$ be the minimal
language 
allowed, namely, the result of starting with $\L(\Scal)$ and then
extending it to include formulas of the
form $\phi \Cond \psi$,
where $\phi$ and $\psi$
are in $\L(\Scal)$.
(Thus, $\Lex(\Scal)$ has nested counterfactuals, while $\L(\Scal)$
does not.)
We take $\LM = \Lex(\Scal)$.

We might hope to identify $\phi \Cond \psi$ with $[\phi]\psi$.
This works fine as long as $\phi$ is a conjunction of primitive
events.  But $[\phi]\psi$ is not defined
in causal models
for an arbitrary Boolean
combination $\phi$ of primitive events.  So, in order to view causal
models as a ccf, we have to define the semantics of $\phi \Cond \psi$
for an arbitrary Boolean combination $\phi$ of primitive events.
%
%
To give semantics to a counterfactual of the form $\phi \Cond \psi \in
\Lex(\Scal)$, we proceed as follows:  Let $\vec{Y}$
be the endogenous variables that appear in $\phi$.  Then we take
$(M,\vec{u}) \models \phi \Cond \psi$ iff there exists a vector $\vec{y}$
of values such that $\phi \land \vec{Y} = \vec{y}$ is consistent
(when viewed as a propositional formula) and
$(M,\vec{u}) \models [\vec{Y} \gets \vec{y}]\psi$.   It is easy to
check that $\phi \Cond \psi$ agrees with $[\phi]\psi$ if $\phi$ is a
conjunction of primitive events, so this is a generalization of the
standard definition of counterfactuals in causal models.

This definition seems somewhat \emph{ad hoc} and, indeed, has some
counter-intuitive properties (see Example~\ref{xam:weird}).  For example, why
restrict $\vec{Y}$ to consisting only of variables that appear in
$\phi$? Indeed, 
there are other definitions of counterfactuals in causal models that
agree with $[\phi]\psi$
if $\phi$ is a conjunction of primitive
propositions. For example, we could take $\phi \Cond \psi$ to be
$[\phi]\psi$ if $\phi$ is a conjunction of primitive propositions, and
$\false$ otherwise.
The current definition has the benefit of allowing us to 
prove that the generalized definition of causality agrees with the HP
definition in causal models (see Theorem~\ref{thm:defsagree}).
\commentout{
However, there is another approach to defining counterfactuals in
causal models that is arguably less \emph{ad hoc} and still gets this
agreement; see the discussion after Theorem~\ref{pro:nobacktracking}.
}

\vspace{-0.1in}
\subsection{Counterfactual structures as a ccf}\label{sec:lewis}

For our purposes, a (Lewis-style) {\em counterfactual structure\/}
$M$ over $\Scal$
is a tuple $(\W,R,\pi)$, where $\W$ is a
finite set of states,
$\pi$ is an \emph{interpretation} that maps each state to a truth
assignment on the primitive events over $\Scal$, and $R$ is a ternary
relation over $\W$.  
Intuitively, $(s,t,u) \in R$ if $t$ is at least as close to $s$ as
$u$ is.\footnote{The presentation here follows that of Halpern
\nciteyear{Hal40}, with some simplifications for ease of
exposition.
Although we focus on Lewis's definition
\cite{Lewis73}, largely the same approach handles that of Stalnaker
\nciteyear{Stalnaker68}, who assumes that there is a unique closest
state, while Lewis allows for a set of closest states.}
Let $t \preceq_s u$ be an abbreviation for $(s,t,u) \in R$.
Define $t \prec_s u$ if $t
 \preceq_s u$ and $u \not\preceq_s t$.  
For simplicity, we 
require that 
$\preceq_s$ be reflexive and transitive and assume that $s \prec_s
t$ for all $t \ne s \in \W$ (so, intuitively, $s$ is the unique state
closest to itself according to $\preceq_s$).
These are standard assumptions in the literature.

As for the language, we again fix a signature $\Scal$.  Taking
$\M^c(\Scal)$ to be the counterfactual structures over signature $\Scal$
(so that the interpretation $\pi$ in these structures gives semantics
to primitive events over $\Scal$), we assume that $\L^{\M^c(\Scal)}$
  includes $\L(\Scal)$ and is closed under conjunction, negation,
  counterfactuals (so that if $\phi, \psi \in \L^{\M^c(\Scal)}$, then
    so is $\phi \Cond \psi$, which means that we have nested counterfactuals).
As usual, given a counterfactual structure $M \in \M^c(\Scal)$ and state $s$, we
define $(M,s) \sat \phi \Cond \psi$ if for all states $t$ such that
$(M,t) \sat \phi$ and there is no state $t'$ such that $(M,t') \sat
\phi$ and $t' \prec_s t$, we also have $(M,t) \sat \psi$.  
That is, $(M,s) \sat \phi \Cond \psi$  if in all the closest states to
$s$ satisfying $\phi$, $\psi$ also holds.  (It follows that 
$(M,s) \sat \phi \Cond \psi$ vacuously if there are no states $t$ such
that $(M,t) \sat \phi$.)

The careful reader will have noticed that this means we are treating
formulas with negations differently in counterfactual structures
and in causal models.  For example, in the
former case, the formula $(X\ne x) \Cond \psi$ is taken to be true at
a state $s$ if 
at the closest state to $s$ such that $X \ne  x$, $\psi$ is true.
  In the latter case, the formula is true if, roughly speaking, there
  exists some value $x'$ such that at the closest state where $X= x'$,
  $\psi$ is true.  If $X$ is a binary variable and takes on only two
  values, say $x$ and $x'$, at the closest state where where $X \ne
  x$, we must have $X=x'$, and the two approaches agree.  But if $X$
  is not binary, the two approaches may not agree.
  As we shall see,
  this distinction disappears given how we deal with causality.

\vspace{-0.1in}
  \section{A more general abstract definition of actual
  causality}\label{sec:abstraction} 
  
We now define actual causality in arbitrary ccfs.
While the focus of this paper is on causal models and counterfactual
structures, we stress that the definition applies far more broadly.  For
example, rather than Lewis-style counterfactual structures, we could
consider Stalnaker-style counterfactual structures \cite{Stalnaker68},
where there is a function $f$ that associates with each state $s$ and formula
$\phi$ a unique state $f(s,\phi)$ closest to $s$ that satisfies
$\phi$.  We can also apply the definition to subclasses of counterfactual
structures (and do), alternate
definitions of causal models, and the extended causal models of Lucas
and Kemp \nciteyear{LK15}.
%
In the definition, the set of possible witnesses is a
parameter.  As we shall see, this provides useful flexibility.

\dfn\label{actcaus}
Given a ccf $\M$ and set $\C_{\X} \subseteq \LM$ of formulas,
$\phi$ is an \emph{actual cause of
$\psi$ with respect to $\C_{\X}$ in 
a model $M \in \M$ and state $s$} if the following
three conditions hold:\index{causal setting}
\begin{description}
\item[{\rm AC1$'$.}]\label{ac1} $(M,s) \sat \phi \land \psi$;
\item[{\rm AC2$'$.}]\label{ac2} There is a
    formula $\tau \in \C_{\X}$ 
   such that $(M,s) \sat \tau$  and
  $(M,s) \sat  (\neg \phi   \land \tau) \Cond  \neg \psi$.
\item[{\rm AC3$'$.}] \label{ac3}\index{AC3}  
 $\phi$ is minimal; there is no formula $\phi' \in \C_{\X}$ such that $\sat \phi
  \rimp \phi'$ and $\not\sat \phi' \rimp \phi$ such that AC2$'$ holds
  with $\phi$ replaced by $\phi'$.
  \bbox
\end{description}
\end{definition}

\vspace{-0.1in}
\subsection{Applying the abstract definition in actual causal models}

This definition is, by design, quite similar to the definition of
actual causality in causal models, but more general.
Causes no longer have to have the form $\vec{X} = \vec{x}$, and we
allow the witnesses in $\vec{W}$ to come from some arbitrary set $\C_{\vec{X}}$ of
formulas that may depend on $\vec{X}$.
But, as we observed above, there is a nontrivial difference in how
negation is dealt with.
As we hinted above, despite this difference, in
causal models, we defined $\Cond$ so that AC2 and AC2$'$ lead to
identical conclusions
(with the appropriate choice of $\C_{\vec{X}}$).

\commentout{
Before proving this result, we prove a result that is interesting in
its own right.  it shows that we can take that the ``witness''
$\vec{x}'$ in AC2 to differ from $\vec{x}$ in all components.

\pro\label{AC2min} To show that $X_1 = x_1 \ldots \land \ldots X_n =
x_n$ is a cause of $\psi$ in a causal setting $(M,\vec{u})$ using
AC1-3, we can assume that the setting $\vec{x}' = (x_1', \ldots,
x_n')$ in AC2 is such that $x_i \ne x_i'$ for $i =1,\ldots, n$.
\epro

\fullv{
\prf Suppose that $\vec{x}' = (x_1', \ldots, x_n')$, and to obtain a
contradiction suppose, without loss
of generality, that $x_1' = x_1$.  Then we can add the conjunct $X_1 =
x_1'$ to $\vec{W} = \vec{w}^*$, and consider the witness
$(x_2', \ldots, x_n')$.  With this change, AC2 continues to hold
(clearly, $(M,\vec{u}) \sat (X_2 = x_2'  \land
\cdots
\land X_n = X_n' \land
(X_1 = x_1' \land \vec{W}= w^*)) \Cond \neg \psi)$, which means that we
have a violation of AC3.
\eprf
}
}

\thm\label{thm:defsagree}
Let $\C_{\vec{X}}$ consist of all formulas $\tau$ such that $\tau$ is a
conjunction of arbitrary (non-negated) primitive events
(so $\C_{\vec{X}}$ is in fact independent of $\vec{X}$).
Then $\vec{X} = \vec{x}$ is a
cause of $\psi$
in a causal setting $(M,\vec{u})$ 
according to AC1-3 iff $\vec{X} = \vec{x}$ is a
cause of $\psi$
with respect to $\C_{\vec{X}}$
in $(M,\vec{u})$ according to AC1$'$-3$'$.
\ethm

The choice of $\C_{\vec{X}}$ in Theorem~\ref{thm:defsagree} 
ensures that,
just as in AC2, in whatever corresponds to the ``closest states $s'$
to $s$  where 
$\vec{X} \ne \vec{x}   \land \tau$ is true, a formula of the form
$\vec{X} = x' \land \tau$ is true, where $\tau$ is the conjunction
of all the primitive events in $\tau$ (but does not involve the
negations of primitive events in $\tau$).  It easily follows that
$(M,s) \sat \tau$, which shows that AC2$'$ ends up with a formula
much like that in AC2.

It is also worth noting  that if $\C'_{\vec{X}} \supseteq \C_{\vec{X}}$, then the forward
direction of Theorem~\ref{thm:defsagree} holds, but the converse may not.
Theorem~\ref{thm:defsagree} does continue to hold if we allow
negated primitive events as conjuncts, but once we allow disjunctions
of primitive events,
the converse fails in general.
For example, if $\tau$ is $\SH=0 \lor \BH=0$ in
Example~\ref{xam:rt},
then
clearly $(M^{rt},U_{11}) \sat \tau$.  Moreover, 
$\BT=0 \land \SH=0 \land \BH=0$ is consistent with $\BT\ne 1 \land \tau$,
and $(M^{rt},U_{11}) \sat [\BT\gets 0, \SH \gets 0, \BH\gets 0](\BS=0)$,
so with this choice of $\C'_{\X}$, $\BT=1$ would be a cause of $\BS=1$.

This example (and Theorem~\ref{thm:defsagree}, for that matter) very
much depend on the fact that $\vec{Y}$ consists precisely of the
variables that appear on the left of the $\Cond$.

\vspace{-0.1in}
\subsection{Applying the abstract definition of causality in recursive
counterfactual structures}

As we now show, the abstract definition of causality, when applied to a
subfamily of counterfactual structures, can be viewed as generalizing the
definition in causal structures.  The subfamily of counterfactual
structures corresponds
in a precise sense to recursive causal models.  Formally,
a counterfactual structure $M'$
\emph{corresponds} to a causal model $M$ if $M$ and $M'$ agree on all
the equations; more precisely, for each endogenous variable $Y$,
if $\W_Y = (\U \union \V)-\{Y\}$, then for each setting $\vec{s}_Y$ of the variables in
  $\W_Y$, and each state $s$, in the closest states to $s$ such that
  $\W_Y = \vec{s}_Y$ holds, we have that $Y =
F_Y(\vec{s}_Y)$.  $M'$
\emph{strongly corresponds} to $M$ if (a) $M'$ corresponds to $M$,
(b) for each assignment $v$ of values to variables, there
is a state in $M'$ where $v$ is the assignment, and
  (c) for each state $s$, if $(M',s) \sat \U = \vec{u}$ and
  $\psi$ is a propositional formula such that $\U = \vec{u} \land
  \psi$ is consistent, then in the closest states to $s$ such that
  $\psi$ holds, we have that $\U = \vec{u}$.%
  \footnote{Although formulas of the form $\U = \vec{u}$ are not in
  the language $\L(\Scal)$, for the purposes of this discussion, we
  assume that we have extended the language to include them,
with the obvious semantics.}
  Roughly speaking, $\U =
  \vec{u}$ continues to hold in the closest states to $s$ unless some
  value in $\U$ is explicitly changed.  $(M',s)$ is (strongly)
  consistent with $(M,\vec{u})$ if
  $M'$ (strongly) corresponds to $M$ and $(M',s) \sat \U = \vec{u}$.
  
    A counterfactual structure is \emph{recursive} if it
        strongly corresponds to a recursive causal model.  In general,
    there may 
    be many recursive counterfactual structures that strongly
    correspond to
a given recursive causal model; the definition of strong
    consistency does not specify how the closest-state relation $R$ in
    the counterfactual structure should work for formulas that are not
    structural equations.
    Nevertheless, as the following result
shows, recursive counterfactual structures are closely related to
(recursive) causal models, at least as far as counterfactual reasoning
is concerned.
We say that $(M',s)$ is a \emph{recursive setting} if $M'$ is a
recursive counterfactual structure and $s$ is a state in $M'$.

%


\pro\label{pro:agree}\cite[Theorem 3.4]{Hal40}  If $(M,\vec{u})$ is a
recursive causal setting and $(M',s)$ is
a recursive setting that is
strongly consistent with
$(M,\vec{u})$, then for
all formulas $\phi \in \L(\Scal)$, $(M,\vec{u}) \sat \phi$  iff
$(M',s) \sat \phi$
(where we identify $\psi \Cond \psi'$ with $[\psi]\psi'$).
\epro


The restriction to $\L(\Scal)$ is critical here, as the following example
shows.

\xam\label{xam:weird} Consider a recursive causal model $M$ with two endogenous
variables, $X$ 
and $Y$, and one exogenous variable $U$, such that $\R(U) = \R(X) =
\R(Y) = \{0,1,2\}$.  Suppose that $U$ is the unique parent of $X$, and
$X$ is the parent of $Y$; the equation for $X$ is $X=U$, and the equation for
$Y$ is $Y=X$.  Let $\psi$ be the formula
$(X \ne 0  \Cond Y=1) \land (X \ne 0 \Cond Y=2)$.
Note that $\psi \notin \L(\Scal)$ due to the antecedent $X \ne 0$.
It is easy to see
that $(M,U=0) \sat \psi$, but there can be no counterfactual structure
$M'$ and state $s$ such that $(M',s) \sat \psi$. \exam

Despite this example, we have the following result (whose proof is
deferred to the appendix).

\thm\label{pro:nobacktracking}
Let $\vec{X}$ be a set of endogenous variables, and
let $\C_{\vec{X}}$ consist of all formulas $\tau$ such that $\tau$ is
a conjunction of (a) arbitrary (non-negated) primitive events, 
and (b) a disjunction of the form $\vec{X} = \vec{x} \lor
\vec{X} = \vec{x}^*$, where $\vec{x}^* \ne \vec{x}$.
If $(M,\vec{u})$ is a recursive causal
setting and  $(M',s)$ is a recursive counterfactual structure that is
strongly consistent with $(M,\vec{u})$ 
then
$\vec{X} = \vec{x}$ is a cause of $\phi$ in
$(M,\vec{u})$ according to AC1-3 iff
$\vec{X} = \vec{x}$ is a cause of $\phi$ with respect to
$\C_{\vec{X}}$ in
$(M',s)$ according to AC1$'$-3$'$.
\ethm

Note that $\C_{\vec{X}}$ differs in Theorems~\ref{thm:defsagree} and
\ref{pro:nobacktracking}.  These results show the power of allowing
$\C_{\vec{X}}$ as a parameter.
\paragraph{On the Role of the Language:}
We emphasize that for Theorem~\ref{pro:nobacktracking} to hold, we
must allow the set
$\C_{\vec{X}}$ of formulas that we condition on to contain
(a quite restricted form of) disjunctions.  If we were to allow arbitrary
disjunctions, in particular, disjunctions that contain $\neg
\phi$, then any formula would become a cause of 
$\phi$, and the definition becomes trivial.
What happens if we restrict
$\C_{\vec{X}}$ to contain only a conjunction of primitive events? Then
our definition and the HP definition would diverge: 
Suppose that
Bob is placed in a room with a ticking bomb, which can be
defused by entering a secret 100-digit combination. Bob can decide to
run away, or try pushing in a number combination. If Bob runs away
and the bomb explodes, should his action of running away be considered
the cause of it? According to the HP definition, it would (since if
Bob had just entered the right combination, then the bomb would be
defused). According to our definition (and disallowing disjunctions
and negations in $\C_{\vec{X}}$), it would depend on what number
combination Bob would enter in the closest world where he does not run
away. For instance, if Bob knows the right combination, then him
running away would be considered a cause, but if he does not,
then it would not.
Arguably, the new definition (without disjunctions) gives the more
reasonable answer in this situation.
On the other hand, if the number combination is short (say,
1 digit), then arguably the HP definition would give the right answer
\fullv{whereas, without disjunctions, ours would not (even if Bob does not
know the secret, assuming that he is allowed multiple trials to enter
it).
This discussion suggests that perhaps allowing a limited number of
disjunctions
may be a reasonable way of defining causality.}
\shortv{whereas, without disjunctions, our new definition would not.}

\vspace{-0.1in}
\section{Backtracking}\label{sec:backtracking}

As has often been observed, when considering the effect of an
intervention such as $X=x$ in a causal model, no variables
``upstream'' of $X$ in the causal network (more precisely, no
variables that are not descendants of $X$) are affected; only
descendants of $X$ can be affected.
By 
way of contrast,
even in recursive counterfactual structures,
in
the state(s) closest to $(M,s)$ where $X=x$ is true, it may well be
the case that ancestors of $X$ have changed their value.
Such backtracking is allowed in Lewis-style models, when considering
closest states. 
%
K\"ugelgen
et al.~\nciteyear{KMB23} introduced a whole mechanism for allowing
backtracking counterfactuals in causal models that involved changing
the values of exogenous variables.  Note that changing the values of
an exogenous variable is not viewed as an intervention; rather, it is
assumed that the agent considering whether $\vec{X} = \vec{x}$ is a
cause of $\phi$ is uncertain about the value of the exogenous
variable, and is considering what would happen to $\phi$ if the value of
the exogenous variable were different.
%
Here we show that our abstract definition of causality allows
backtracking in a natural way, without having to introduce a special
mechanism for it.  Indeed, we can move smoothly from allowing
backtracking to forbidding it, just by fixing an appropriate set 
of variables.

It follows from Proposition~\ref{pro:nobacktracking} that we have no
backtracking in recursive counterfactual structures that strongly
correspond to causal models.  But once we allow  the context to
change when considering causality, we can get backtracking.
This is exactly what happens when we move from 
counterfactual structures that strongly correspond to a causal model
to ones that just correspond to a causal model.

\xam\label{xam:SuzyLewis}
Recall the causal setting $(M^{rt},U=11)$ for Example~\ref{xam:rt},
the rock-throwing example.  Let $M' = (\W,R,\pi)$ be consistent 
with $M^{rt}$.  Let $s$ be a state such that $(M',s)$ is consistent with 
$(M^{rt},U=11)$; in particular, $(M',s) \sat \ST = \BT = \SH = \BS
= 1 \land \BH = 0$.  Define $R$ so that the closest state $s'$ to $s$
is such that $(M',s') \sat U=00$, so that $(M',s') \sat 
\ST = \BT = \SH = \BH = \BS = 0$.  With these choices, $M'$
does not strongly correspond to $M$.  It is easy to see that
$(M',s) \sat \BT 
= 0 \Cond \BS =0$, so $\BT=1$ is a cause of $\BS=1$ in $(M',s)$.  By
allowing the context to change in the closest state to $s$, we get
backtracking.
\exam

By working with  a counterfactual setting $(M',s)$
that corresponds to a causal
setting  $(M,\vec{u})$, we can capture causality with and without
backtracking in $(M,\vec{u})$, simply by adding the appropriate conjuncts to
the formula $\tau$ in AC2$'$.  For example, if we want capture why
$\vec{X}=\vec{x}$ is a cause of $\phi$
without backtracking, we simply add the conjunct $\U = \vec{u}$ to
$\psi$.  If we want to allow some backtracking, we may want to fix the
values of some ancestors of $\vec{X}$ to their actual value, without
fixing $\U$.
This makes it clear that whether or not we have backtracking depends
on the set of formulas $\C_{\X}$ from which $\tau$ is drawn.

\vspace{-0.1in}
\section{An Abstract  Definition of Explanation}\label{sec:explanation1}
Our approach can be extended to deal with causal explanation, using
the HP definition \cite{Hal48,HP01a}.
We start by reviewing the HP definition of explanation; for more
details and intuition, see \cite{Hal48}.
As noted in the introduction, the HP definition of causality is
relative to an agent and, specifically, what the agent knows.  In causal
models, an agent's knowledge is captured by the agent's  \emph{epistemic state}
$\K$, where $\K$ is a set of contexts settings with a fixed causal model
$M$.%
\footnote{In \cite{Hal48} there is also assumed to be a probability on
$\K$; for simplicity, we ignore the probability here. Also, in
\cite{Hal48}, an agent may be uncertain about the model as well as the
context.  Assuming that the model is fixed, as we are doing here, is
without loss of 
generality.  If an agent is uncertain about the model, that
uncertainty can be encoded into the context by having a special
exogenous variable $u^*$ that determines the model; that is, the value
of $u^*$ gives the model.  Given a model $M$,
the equations in $M$ can be encoded using equations that hold when $u^*=M$.}
Intuitively, the contexts in $\K$ are the ones that the agent
considers possible, and reflects the agent's uncertainty regarding how
the world works (represented by the equations involving context
$\vec{u}$) and what
is currently true (represented by the context $\vec{u}$).
As is standard, we say that an agent \emph{knows} $\phi$ if $\phi$ is
true at all the settings in $\K$.

We next give the formal definition of explanation in causal models,
and then give intuition for the clauses, particularly EX1(a).
\dfn\label{def:explanation1} 
{\em $\vec{X} = \vec{x}$ is an explanation
of $\psi$ in a model $M$ relative to a set $\K$ of contexts in $M$\/} 
if the following conditions hold:
\begin{description}
\item[{\rm EX1(a).}] 
If $\vec{u} \in \K$ and $(M,\vec{u}) \sat \vec{X} = \vec{x} \land
\psi$, then 
  there exists a conjunct $X=x$ of \mbox{$\vec{X} = \vec{x}$} and a (possibly empty)
  conjunction $\vec{Y} = 
  \vec{y}$ such that \mbox{$X = x \land \vec{Y} = \vec{y}$} is
  a cause of $\psi$ in $(M,\vec{u})$.  
\item[{\rm EX1(b).}]  $(M,\vec{u}') \sat [\vec{X} \gets \vec{x}]\psi$ for all
  contexts $\vec{u}' \in \K$.  
\item[{\rm EX2.}] $\vec{X}$ is minimal; there is no strict subset
  $\vec{X}'$ of $\vec{X}$ such that $\vec{X}' = \vec{x}'$ satisfies
  EX1(a) and EX1(b), where $\vec{x}'$ is the restriction of
$\vec{x}$ to the variables in $\vec{X}$.   
\item[{\rm EX3.}]
For some $\vec{u} \in \K$, we have that $(M,\vec{u})) \sat \vec{X}
= \vec{x} \land \psi$.  
(The agent considers possible a setting 
where the explanation and explanandum both hold.) 
\end{description}

The explanation is \emph{nontrivial} if it also satisfies 
\begin{description}
\item[{\rm EX4.}]
  $(M,\vec{u}') \sat \neg(\vec{X}=\vec{x}) \land \psi$ for some $\vec{u}' \in
  \K$
(The explanation is not already known given the observation of
  $\psi$.)
\eprf
\end{description}
\end{definition}

The key part of the definition is EX1(b).  Roughly speaking, it says
that the explanation $\vec{X} = \vec{x}$ is a \emph{sufficient cause}
for $\psi$: for all settings that the agent considers possible,
intervening to set $\vec{X}$ to $\vec{x}$ results in $\psi$.  (See
\cite[Chapter 2.6]{Hal48} for a formal definition of sufficient cause.)
EX2 and EX3 should be fairly clear.  
EX4 is meant to capture the intuition (discussed in the introduction)
that what counts as an explanation depends on what the agent knows.
In the formal model, this is captured by the set $\K$.  
That leaves EX1(a).
Roughly speaking, it says that the explanation causes the
explanandum.  But there is a tension between EX1(a) and EX1(b) here:
we may need to add conjuncts to the explanation to ensure that it
suffices to make $\psi$ true in all contexts (as required by EX1(b)).
But these extra conjuncts may not be necessary to get causality in all contexts.
At least one of the conjuncts of $\vec{X} = \vec{x}$ must be
part of a cause of $\psi$, but the cause can
include extra conjuncts.  To understand why, it is perhaps best to
look at an example.

\begin{example}
Consider a version of the rock-throwing example where now in the
causal model $M$ it is
possible that (a) Suzy's rock gets to the bottle first, (b) Billy's rock
gets to the bottle first, or (c) both rocks arrive simultaneously.  We
capture this using 12 contexts of the form $u_{ijk}$ where, as before
$i=1$ if Suzy throws and $j=1$ if Billy throws.  If $k=1$ (resp.,
$k=2$, $k=3$) and both Billy and Suzy throw, then Suzy's rock gets to
the bottle first (resp., Billy's rock gets to the bottle first; the
rocks arrive simultaneously).  In all cases, just one rock hitting the
bottle suffices for the bottle to shatter.

Consider $\K_1 = \{u_{111}, u_{112}, u_{101}\}$.
An explanation for the bottle shattering relative to $\K_1$ is
$\ST = 1 \land \BT=1$.  Each of $\ST=1$ and $\BT=1$ satisfies EX1(b).
But while $\ST=1$ is a cause of the bottle shattering in
$(M,\vec{u}_{111})$ and $(M,\vec{u}_{101})$, it is not a cause of the
bottle shattering in 
$(M,\vec{u}_{111})$, while $\BT=1$ is. On the other hand, $\BT=1$ is not a
cause of the bottle shattering in $(M,\vec{u}_{112})$.
Thus, EX1(a) holds for 
$\ST = 1 \land \BT=1$, taking $\ST=1$ to be the relevant conjunct in 
$(M,\vec{u}_{111})$ and $(M,\vec{u}_{111})$ and $\BT=1$ to be the
relevant conjunct in  
$(M,\vec{u}_{111})$. (Note that we do not have to take $u_{101}$ into
account for EX1(a), because $(M,\vec{u}_{111}) \not\sat \ST = 1 \land
\BT=1$.)  It is easy to see that EX2, EX3, and EX4 hold in this case.

Next consider $\K_2 = \{u_{111}, u_{112}\}$.  In this case, $\ST=1
\land \BT=1$ is an explanation of the bottle shattering relative to
$\K_2$, but a trivial one; it was already known. 

Finally, consider $\K_3 = \{u_{003}, u_{103}, u_{013}, u_{113}\}$.  In
this case, each of $\ST=1$
and $\BT=1$ is an explanation of the bottle shattering relative to $\K_3$.
However, note that in the context $(M,u_{113})$,
$\ST=1$ is \emph{not} a cause of the bottle shattering; if Suzy
doesn't throw, then the bottle still shatters (since Billy's rock hit
the bottle).  However, $\ST=1 \land \BT=1$ is a cause, and EX1(a)
allows adding a conjunct $Y=y$ (in this case, $\BT=1$) to get a cause.
Note that $\ST=1 \land \BT=1$ is not an explanation of the bottle
shattering; it violates the minimality condition EX2.
\exam

We now show how to translate each of these conditions to our more
abstract framework.  To start with, the obvious analogue of $\K$ is
just a set of worlds in some model in a ccf.

\dfn\label{explanation2}
Given a ccf $\M$ and set $\C_{\X} \subseteq \LM$ of formulas,
$\phi$ is an \emph{explanation of
$\psi$ with respect to $\C_{\X}$ in 
a model $M \in \M$ and state $s$ relative to a set $\K$ of states in
$M$} if the following 
three conditions hold:
\begin{description}
\item[{\rm EX1(a)$'$.}] If $s \in \K$ and $(M,s) \sat \phi \land
  \psi$, then there exist formulas $\tau_1, \tau_2 \in \C_{\X}$
  such that (i) $\tau_1$ is not valid, (ii) $\sat \phi \rimp \tau_1$,
  (iii) $\sat \tau_2 \rimp \tau_1$, and (iv) $\tau_2$ is a cause of $\psi$
with respect to $\C_{\X}$ in $(M,s)$ (according to Definition~\ref{actcaus}).
\item[{\rm EX1(b)$'$.}] $(M,s') \sat \phi \Cond \psi$ for
  all states $s' \in \K$.
\item[{\rm EX2$'$.}]   $\phi$ is minimal; there is no formula $\phi'$ such that $\sat \phi
  \rimp \phi'$ and $\not\sat \phi' \rimp \phi$ such that EX1(a) and
  EX1(b) hold   with $\phi$ replaced by $\phi'$.
\item[{\rm EX3$'$.}]  For some $s \in \K$, we have that $(M,s)
  \sat \phi \land \psi$.
  \end{description} 
Again, the explanation is nontrivial if
\begin{description}
\item[{\rm EX4$'$.}]   $(M,s') \sat \neg \phi \land \psi$ for some $s' \in
  \K$.
  \bbox
  \end{description} 
\end{definition}

All the primed conditions except for EX1(a)$'$ are fairly direct
translations of their unprimed counterparts.  For EX1(a)$'$, note that
the effect of adding a conjunction $\vec{Y} = \vec{y}$ to $X=x$ in
EX1(a) is to get a formula that implies $X=x$; that is why we have
$\sat \tau_2 \rimp \tau_1$ in part (iii) or EX1(a)$'$.  Similarly, the
conjunct $X=x$ in EX1(a) (which corresponds to $\tau_1$ in EX1(a)$'$)
is implied by $\vec{X} = \vec{x}$ (which correspond to $\phi$ in
EX1(a)$'$).  To understand the requirement that $\tau_1$ not be valid,
note that if it were valid, then conditions (ii) and (iii) would hold
vacuously, so (iv) would hold as long as we can find a formula that
causes $\psi$ with respect to $\C_{\X}$ in $(M,s)$.  We want $\tau_1$
to have a nontrivial connection to $\phi$.  In EX1(a), this is
accomplished by making it a conjunct of $\phi$.  In EX1(a)$'$, this is
accomplished by having it be a nonvalid formula that is implied by
$\phi$.

In any case, we now get the analogue of
Theorem~\ref{thm:defsagree}. showing that the abstract definition of
explanation generalizes the original HP definition.

\thm\label{thm:defsagree1}
Let $\C_{\vec{X}}$ consist of all formulas $\tau$ such that $\tau$ is a
conjunction of arbitrary (non-negated) primitive events, and let $\K$
be a set of contexts in a causal model $M$.
Then $\vec{X} = \vec{x}$ is an 
explanation of $\psi$ in a causal setting $(M,\vec{u})$ relative to $\K$
according to EX1--3 (resp., EX1--4) iff $\vec{X} = \vec{x}$ is an
explanation of $\psi$
with respect to $\C_{\vec{X}}$ relative to $\K$
in $(M,\vec{u})$ according to EX1$'$-3$'$ (resp., EX1$'$-4$'$).
\ethm

We also get an analogue of Theorem~\ref{pro:nobacktracking}, but to get
this we need an even tighter connection
between
recursive counterfactual
structures and recursive causal models.   A recursive causal model $M$
and a recursive counterfactual structure $M'$ are \emph{compatible} if
for every context $\vec{u}$ in $M$ there is a state $s$ in $M'$ such
that $(M,\vec{u})$ is strongly consistent with $(M,s)$, and for every
state $s$ in $M'$ there is a context $\vec{u}$ in $M$ such that
$(M,\vec{u})$ is strongly consistent with $(M,s)$.  A set $\K$ of
contexts in $M$ and a set $\K'$ of states in $M'$ are
\emph{compatible} if 
for every context $\vec{u}$ in $\K$ there is a state $s$ in $\K'$ such
that $(M,\vec{u})$ is strongly consistent with $(M,s)$, and for every
state $s$ in $\K'$ there is a context $\vec{u}$ in $\K$ such that
$(M,\vec{u})$ is strongly consistent with $(M,s)$.  
Finally, a structure $M$ is \emph{acceptable} if $(M,s)$ is acceptable
for all states $s$ in $M$.

\thm\label{pro:nobacktracking1}
Let $\vec{X}$ be a set of endogenous variables, and
let $\C_{\vec{X}}$ consist of all formulas $\tau$ such that $\tau$ is
a conjunction of (a) arbitrary (non-negated) primitive events, 
and (b) a disjunction of the form $\vec{X} = \vec{x} \lor
\vec{X} = \vec{x}^*$, where $\vec{x}^* \ne \vec{x}$.
Given a recursive causal model $M$ and a recursive counterfactual
structure $M'$ such that $M$ and $M'$ are compatible, and compatible
sets $\K$  of contexts in $M$ and $\K'$ of states in $M'$,
if $(M,\vec{u})$ is a recursive causal
setting and  $(M',s)$ is strongly consistent with $(M,\vec{u})$, then 
$\vec{X} = \vec{x}$ is an explanation of $\psi$ relative to $\K$ in
$(M,\vec{u})$ 
according to EX1--3 (resp., EX1-4) iff
$\vec{X} = \vec{x}$ is an explanation of $\psi$ with respect to
$\C_{\vec{X}}$ relative to $\K'$ in
$(M',s)$ according to EX1$'$-3$'$ (resp., EX1$'$-4$'$).
\ethm

\vspace{-0.2in}
\shortv{
\section{Conclusions}\label{sec:conclusions}
We have given an abstract definition of causality that applies to
arbitrary ccfs.  In particular, it
applies to causal models (where it agrees with the standard
HP definition) and to Lewis-style counterfactual structures
(where it generalizes the HP definition).  As we have
observed, having 
such a general definition lets us apply the definition far more
broadly, and to richer, more expressive languages.

Roughly speaking, the abstract definition says that $A$ is a cause of
$B$ if, conditional on $C$, $A$ is a but-for cause of $B$.  We have
restricted $C$ to come from some set that we denoted $\C_{\vec{X}}$.  We are
currently trying to  understand the impact of $\C_{\vec{X}}$ on the definition.
While in causal models it sufficed to require that $\C_{\vec{X}}$ be a
conjunction of primitive events to capture the HP notion of causality,
in counterfactual structures we also required $\C_{\vec{X}}$ to
contain (restricted) disjunctions. What happens if we were to restrict
$\C_{\vec{X}}$ to contain only a conjunction of primitive events? At
this point, the two definitions would diverge: Consider a scenario
where Bob is placed in a room with a ticking bomb, which can be
defused by entering a secret 100-digit combination. Bob can decide to
run away, or try pushing in a number combination. If Bob runs away
and the bomb explodes, should his action of running away be considered
the cause of it? According to the HP definition, it would (since if
Bob had just entered the right combination, then the bomb would be
defused). According to our definition (and disallowing disjunctions
and negations in $\C_{\vec{X}}$), it would depend on what number
combination Bob would enter in the closest world where he does not run
away. For instance, if Bob knows the right combination, then him
running away would be considered a cause, but if he does not,
then it would not.
Arguably, the new definition (without disjunctions) gives the more
reasonable answer in this situation.
On the other hand, if the number combination is short (say
1 digit), then arguably the HP definition would give the right answer
whereas, without disjunctions, ours would not (even if Bob does not
know the secret, assuming that he is allowed multiple trials to enter
it).

We can, to some extent, deal with the problem by considering
\emph{normality} \cite{Hal48}, where, intuitively, we want 
to consider $\vec{X} = \vec{x}$ a cause of $\phi$ only if the formula
$\vec{X} = \vec{x}' \land \vec{W} = \vec{w}^*$ in AC2 is not abnormal.
As argued in \cite{Hal48}, this restriction leads to an arguably more
reasonable notion of causality.  We are currently exploring the
possibility of capturing normality by allowing disjunctions in $\C_X$,
but restricting all the disjuncts to be normal, in some appropriate
sense,
or by restricting to formulas using a small number of disjunctions.
Again, this would show that by conditioning on the appropriate set
of formulas, we can capture important intuitions.
}

\appendix
\vspace{-0.2in}
\section{Proofs}
For the reader's convenience, we repeat the statement of the results.

\smallskip

\othm{thm:defsagree}
Let $\C_{\vec{X}}$ consist of all formulas $\tau$ such that $\tau$ is a
conjunction of arbitrary (non-negated) primitive events
(so $\C_{\vec{X}}$ is in fact independent of $\vec{X}$).
Then $\vec{X} = \vec{x}$ is a
cause of $\psi$
in a causal setting $(M,\vec{u})$ 
according to AC1-3 iff $\vec{X} = \vec{x}$ is a
cause of $\psi$
with respect to $\C_{\vec{X}}$
in $(M,\vec{u})$ according to AC1$'$-3$'$.
\eothm

\smallskip

\prf Suppose that $\vec{X} = \vec{x}$ is a
cause of $\psi$
in $(M,\vec{u})$
according to AC1-3.  Clearly AC1$'$ and AC3$'$ are
essentially equivalent to AC1 and AC3, so to show that $\vec{X}
= \vec{x}$ is a 
cause of $\psi$
with respect to $\C_{\vec{X}}$
according to AC1$'$-3$'$, it suffices to show that
AC2$'$ holds.

Since AC2 holds, there exist a set $\vec{W}$ of endogenous variables,
$\vec{w}^* \in \R(\vec{W}$), and 
$\vec{x}' \in \R(\vec{X})$ such that $(M,\vec{u})
\sat \vec{W} = \vec{w}^*$ and
$
(M,\vec{u}) \sat [\vec{X} \gets \vec{x}', \vec{W} \gets
    \vec{w}^*]\neg \psi.$
We must have $\vec{x}' \ne \vec{x}$, since $(M,\vec{u}) \sat
[\vec{X} \gets \vec{x}, \vec{W} \gets \vec{w}^*]\psi$.
    Let $\tau$ be $\vec{W} = \vec{w}^*$.
  Clearly, $\tau \in \C_{\vec{X}}$;
by assumption,
    $(M,\vec{u}) \sat \tau$.
Moreover, $\vec{X} = \vec{x}' \land \vec{W} = \vec{w}^*$ is consistent with
        $\vec{X} \ne \vec{x} \land \tau$ (since $\vec{W}$ and
    $\vec{X}$ are disjoint).
    Let $\vec{Y}$ consist of the endogenous variables in $\neg(\vec{X} =
    x) \land \tau$; note that $\vec{Y} = \vec{W} \union \vec{X}$.
Choose $\vec{y} \in \R(\vec{Y})$ so that $\vec{Y}= \vec{y}$ is
    $\vec{X} = \vec{x}' \land \vec{W} = \vec{w}^*$.  Thus,
$\vec{X} \ne \vec{x} \land \tau \land \vec{Y} = \vec{y}$ is
consistent.    By assumption, $(M,\vec{u}) \sat [\vec{Y} =
\vec{y}]\neg \psi$.
By definition, $(M,\vec{u}) \sat
(\vec{X} \ne \vec{x} \land \tau) \Cond \neg \psi$, so AC2$'$
holds, as desired.

For the converse, suppose that $\vec{X} = \vec{x}$ is a
cause of $\phi$
with respect to $\C_{\vec{X}} $
in $(M,\vec{u})$ according to AC1$'$-3$'$.  
Again, it suffices to show that
AC2 holds in $(M,\vec{u})$.   Since AC2$'$ holds, there must exist
a formula $\tau
\in \C_{\vec{X}}$
such that $(M,\vec{u}) \sat \tau$ and
$(M,\vec{u}) \sat (\vec{X} \ne \vec{x} \land \tau) \Cond \neg \psi.$  Let
$\vec{Y}$ be the endogenous variables that appear in
$\vec{X} \ne \vec{x} \land \tau$.  By definition, there must be a
vector $\vec{y}$ of values such that
$\tau \land (\vec{Y} = \vec{y})$ is consistent and
$(M,\vec{u} \sat [\vec{Y} = \vec{y}] \neg \psi.$  $\vec{Y}$ must
include $\vec{X}$. Let $\vec{x}'$ be the restriction of $\vec{y}$ to
the variables in $\vec{X}$.  

For each variable $Y \in \vec{Y}'  = \vec{Y} - \vec{X}$, there
is a value $y$ such that $Y=y$ is a conjunct of $\tau$; $Y=y$ must
also be a conjunct of $\vec{Y} = \vec{y}$, since $\vec{Y}=\vec{y}$ is
consistent with $\tau$.  Thus, $(M,\vec{u}) \sat Y=y$ and, more
generally, if $\vec{y}'$ is 
the restriction of the values in $\vec{y}$ to the variables in
$\vec{Y}'$, we have that $(M,\vec{u}) \sat \vec{Y}' = \vec{y}'$.  
It follows that AC2 holds, taking $\vec{W} = \vec{Y}'$ and $\vec{x}'$
as defined above.
\eprf

\othm{pro:nobacktracking}
Let
$\C_{\vec{X}}$ consist of all formulas $\tau$ such that $\tau$ is
a conjunction of (a) arbitrary (non-negated) primitive events, 
and (b) a disjunction of the form $\vec{X} = \vec{x} \lor
\vec{X} = \vec{x}^*$, where $\vec{x}^* \ne \vec{x}$.
If $(M,\vec{u})$ is a recursive causal
setting and  $(M',s)$ is strongly consistent with 
$(M,\vec{u})$, then
$\vec{X} = \vec{x}$ is a cause of $\phi$ in
$(M,\vec{u})$ according to AC1-3 iff
$\vec{X} = \vec{x}$ is a cause of $\phi$ with respect to
$\C_{\vec{X}}$ in
$(M',s)$ according to AC1$'$-3$'$.
\eothm

\smallskip

\prf Suppose that $\vec{X} = \vec{x}$ is a cause of $\phi$ in
a causal setting 
$(M,\vec{u})$ (according to AC1-3), and $(M',s)$ is a recursive
counterfactual structure strongly consistent with $(M,\vec{u})$.
We want to show that $\vec{X} = \vec{x}$ is a cause of $\phi$ in
$(M',s)$.  The argument is similar in spirit to the first half of
the proof of Theorem~\ref{thm:defsagree}.

Again, AC1$'$
  and AC3$'$ are immediate; we need to show that AC2$'$ holds.  Since
  AC2 holds, there is a set $\vec{W}$ of endogenous variables, a
  setting $\vec{w}^*$ such that $(M,\vec{u}) \sat \vec{W} =
  \vec{w}^*$, and a
  setting $\vec{x}'$ of the variables in $\vec{X}$ such that 
  $(M,\vec{u}) \sat [\vec{X} \gets \vec{x}', \vec{W} \gets {w}^*]\neg \phi$.
  Let $\tau$ be $\vec{W} = \vec{w}^* \land (\vec{X} = \vec{x} \lor
  \vec{X} = \vec{x}')$.  Clearly, $\tau \in \C_{\vec{X}}$
and $(M,\vec{u})) \sat \tau$.  Since $(M',s)$ is strongly consistent with
$(M,\vec{u})$,
by Proposition \ref{pro:agree},
we have that $(M',s) \sat \tau$.
  Moreover,
    $(M',s) \sat
    (\vec{X} \ne \vec{x} \land \tau) \Cond \neg \phi$.
    The result
  follows. 

  For the converse, suppose that $\vec{X} = \vec{x}$ is a
cause of $\phi$ with respect to $\C_{\vec{X}} $ in $(M',s)$.  To show
that $\vec{X} = \vec{x}$ is a cause of $\phi$ in $(M,\vec{u})$, yet
again, it suffices to show that AC2 holds.  Since AC2$'$ holds,
there must exist a formula $\tau
\in \C_{\vec{X}}$
such that $(M',s) \sat \tau$,
and $(M',s) \sat (\vec{X} \ne\vec{x} \land
\tau) \Cond \neg \psi$.
There is a conjunction of the form $\tau' \land \vec{X} = \vec{x}^*$
of non-negated primitive events that is 
equivalent to $\vec{X} \ne \vec{x} \land \tau$, where $\tau'$ does
not mention any variable in $\vec{X}'$.
Thus, $(M',s) \sat (\vec{X} =  \vec{x}^* \land \tau') \Cond \neg \psi$.
By 
Proposition~\ref{pro:agree}, it follows that $(M,\vec{u}) \sat 
[\vec{X} =  \vec{x}^* \land \tau'] \neg \psi$.
Since $(M',s) \sat \tau$, we must have $(M',s)
\sat \tau'$, so by Proposition~\ref{pro:agree}, $(M,\vec{u}) \sat
\tau'$.  Thus, taking $\vec{W} = \vec{w}^*$  in AC2 to be $\tau'$,
it follows that AC2 holds, as desired.
\eprf

\vspace{-0.2in}
\paragraph{Acknowledgments:}
Joe Halpern's work was suppprted in part by 
NSF grant FMitF-2319186,
ARO grant W911NF-17-1-0061,  MURI grant W911NF-19-1-0217
from the ARO, and a grant from Open Philanthropy.
Rafael Pass's work was supported in part by 
NSF Award CNS 2149305, AFOSR Award FA9550-23-1-0387, AFOSR Award
FA9550-24-1-0267 and ERC Advanced Grant KolmoCrypt - 1 01142322. 
Any opinions, findings, conclusions, or recommendations expressed in
this material are those of the author(s) and do not necessarily
reflect the views of
the funders.

\bibliographystyle{eptcs}

\end{document}